\documentclass[11pt,a4paper]{article}
\usepackage[hyperref]{emnlp2020}
\usepackage{times}
\usepackage{xspace}
\usepackage{latexsym}
\usepackage{fontawesome}
\usepackage{multirow}
\usepackage{graphicx}
\usepackage{subcaption}

\newcommand\BLEU{\textsc{Bleu}\xspace}
\newcommand\BLEUp{\textsc{BleuP}\xspace}

\usepackage{microtype}

\aclfinalcopy 

\title{Human-Paraphrased References Improve Neural Machine Translation}

\author{Markus Freitag, George Foster, David Grangier, Colin Cherry\\
  Google Research \\
  {\tt \{freitag,fosterg,grangier,colincherry\}@google.com}}

\date{}

\begin{document}
\maketitle

\begin{abstract}

Automatic evaluation comparing candidate translations to human-generated paraphrases of reference translations has recently been proposed by \newcite{freitag2020bleu}. When used in place of original references, the paraphrased versions produce metric scores that correlate better with human judgment. This effect holds for a variety of different automatic metrics, and tends to favor natural formulations over more literal ({\em translationese}) ones. In this paper we compare the results of performing end-to-end system development using standard and paraphrased references. With state-of-the-art English-German NMT components, we show that tuning to paraphrased references produces a system that is significantly better according to human judgment, but 5 BLEU points worse when tested on standard references. Our work confirms the finding that paraphrased references yield metric scores that correlate better with human judgment, and demonstrates for the first time that using these scores for system development can lead to significant improvements.

\end{abstract}

\section{Introduction}

Machine Translation (MT) has shown impressive progress in recent years.
Neural architectures~\cite{bahdanau2014attention,gehring-etal-2017-convolutional,vaswani2017attention} have greatly contributed to this 
improvement, especially for languages with abundant training data~\cite{bojar2016wmt,bojar2018wmt,barrault2019wmt}. 
This progress creates novel challenges for the evaluation of machine translation, 
both for human~\cite{toral2020reassessing,laubli2020recommendations} and automated evaluation 
protocols~\cite{lo2019yisi,zhang19bertscore}.

Both types of evaluation play an important role in machine translation~\cite{koehn2010book}. While human evaluations provide a gold standard evaluation, they involve a fair amount of careful and hence expensive work by human assessors. Cost therefore limits the scale of their application. On the other hand, automated evaluations are much less expensive. They typically only involve human labor when collecting human reference translations and can hence be run at scale to compare a wide range of systems or validate design decisions. The value of automatic evaluations 
therefore resides in their capacity to be used as a proxy for human evaluations for large scale comparisons and system development.

The recent progress in MT has raised concerns about whether automated evaluation methodologies reliably reflect human ratings in high accuracy ranges. In particular, it has been observed that the best systems according to humans might fare less well with automated metrics~\cite{barrault2019wmt}. Most metrics such as \BLEU~\cite{papineni2002bleu} and TER~\cite{snover2006study} measure overlap between a system output and a human reference translation. More refined ways to compute such overlap have consequently been proposed~\cite{banerjee2005meteor,lo2019yisi,zhang19bertscore}.

Orthogonal to the work of building improved metrics, \newcite{freitag2020bleu} hypothesized that human references are also an important factor in the reliability of automated evaluations. In particular, they observed that standard references exhibit simple, monotonic language due to human 
`translationese` effects. These standard references might favor systems which excel at reproducing these effects, independent of the underlying translation quality. They showed that better correlation between human and automated evaluations could be obtained when replacing standard references with {\it paraphrased} references, even when still using surface overlap metrics such as BLEU~\citep{papineni2002bleu}. The novel references, collected by asking linguists to paraphrase standard references, were shown to steer evaluation away from rewarding translation artifacts. This improves the assessment of alternative, but equally good translations.

Our work builds on the success of paraphrased translations for evaluating  existing systems, and asks if different design choices could have been made when designing a system with such an evaluation protocol in mind. This examination has several potential benefits: it can help identify choices which improve BLEU on standard references but have limited impact on final human evaluations; or those that result in better translations for the human reader, but worse in terms of standard reference BLEU. Conversely, it might turn out that paraphrased references are not robust enough to support system development due to the presence of `metric honeypots': settings that produce poor translations, but which are nevertheless assigned high BLEU scores.

To address these points, we revisit the major design choices of the best English$\to$German system from WMT2019~\cite{ng-EtAl:2019:WMT} step-by-step, and measure their impact on standard reference BLEU as well as on paraphrased BLEU. This allows us to measure the extent to which steps such as data cleaning, back-translation, fine-tuning, ensemble decoding and reranking benefit standard reference BLEU more than paraphrase BLEU. Revisiting these development choices with the two metrics results in two systems with quite different behaviors. We conduct a human evaluation for adequacy and fluency to assess the overall impact of designing a system using paraphrased BLEU.

Our main findings show that optimizing for paraphrased BLEU is advantageous for human evaluation when compared to an identical system optimized for standard BLEU. The system optimized for paraphrased BLEU significantly improves WMT newstest19 adequacy ratings (4.72 vs 4.27 on a six-point scale) and fluency ratings (63.8\% vs 27.2\% on side-by-side preference) despite scoring 5 BLEU points lower on standard references.

\section{Related Work}
\label{sec:related_work}

Collecting human paraphrases of existing references has recently been shown to be useful for system evaluation~\cite{freitag2020bleu}.
Our work considers applying the same methodology for system tuning. There is some earlier work relying on {\it automated} paraphrases 
for system tuning, especially for Statistical Machine Translation (SMT). \newcite{madnani2007paraphrase} introduced an automatic paraphrasing technique based on English-to-English translation of full sentences using a statistical MT system, and showed that this permitted reliable system tuning using half as much data. Similar automatic paraphrasing has also been used to augment training data, e.g.~\cite{marton2009paraphrase}, but relying on standard references for evaluation. In contrast to human paraphrases, the quality of current machine generated paraphrases degrades significantly as overlap with the input decreases~\cite{mallinson-2017-paraphrasing,aurko2019paraphrases}. This makes their use difficult for evaluation since~\cite{freitag2020bleu} suggests that substantial paraphrasing -- `paraphrase as much as possible` -- is necessary for evaluation.

Our work can be seen as replacing the regular BLEU metric with a new paraphrase BLEU metric for system tuning. Different alternative automatic evaluation metric have also been considered for system tuning \cite{he2010metric,servan2011optimising} with Minimum Error Rate Training, MERT \cite{och2003minimum}. This work showed some specific cases where Translation Error Rate (TER) was superior to \BLEU.

Our work is also related to the bias that the human translation process introduces in the references, including source language artifacts---{\it Translationese}~\cite{Koppel:2011:TD:2002472.2002636}---as well as source-independent artifacts---{\it Translation Universals}~\cite{mauranen2004translation}. The professional translation community studies both systematic biases inherent to translated texts \cite{baker1993corpus, selinker1972interlanguage}, as well as biases resulting specifically from interference from the source text \cite{toury1995descriptive}. For MT, \citet{Freitag19} point at Translationese as a source of mismatch between BLEU and human evaluation, raising concerns that overlap-based metrics might reward hypotheses with translationese language more than hypotheses using more natural language. The impact of Translationese on human evaluation of MT has recently received attention as well~\citep{Toral18,Zhang19,Graham19}. More generally, the question of bias to a specific reference has also been raised, in the case of monolingual manual evaluation
\citep{fomicheva-specia-2016-reference,ma-etal-2017-investigation}. Different from the impact of Translationese on evaluation, the impact of Translationese in the training data has also been studied~\cite{kurokawa09,Lembersky12adapting,bogoychev2019domain,riley-etal-2020-translationese}.

Finally, our work is also related to studies measuring the importance of the test data quality, looking specifically at the test set translation direction. For SMT evaluation, \newcite{lembersky2012language} and \newcite{stymne2017effect} explored how the translation direction affects translation results. \newcite{holmqvist2009improving} noted that the original language of the test sentences influences the BLEU score of translations. They showed that the BLEU scores for target-original sentences are on average higher than sentences that have their original source in a different language. Recently, a similar study was conducted for neural MT~\cite{bogoychev2019domain}.

\section{Experimental Setup}

We first describe data and models, then present our human evaluation protocol.

\subsection{Data}
\label{sec:data}

We ran all experiments on the WMT 2019 English$\to$German news translation task~\citep{barrault2019wmt}. The task provides $\sim$38M parallel sentences. As German monolingual data, we concatenate all News Crawl data from 2007 to 2018, comprising $\sim$264M sentences after removing duplicates.

\begin{figure}[ht]
    \centering
    \begin{subfigure}[b]{0.22\textwidth}
        \includegraphics[width=\textwidth]{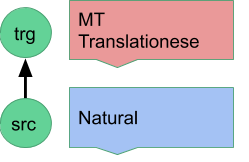}
        \caption{Forward-translated, i.e. source original}
    \end{subfigure}
    \quad
    \begin{subfigure}[b]{0.22\textwidth}
        \includegraphics[width=\textwidth]{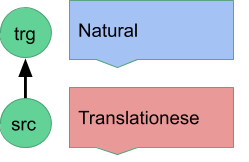}
        \caption{Backward-translated, i.e. target original}
    \end{subfigure}    
    \caption{Sentences in a test set are either natural in the source and forward-translated into the target language, or vice-versa. If a test set consists of both kinds of sentences, we call it a joint test set. WMT English$\to$German newstest2018 is a joint test set with half of the sentences being forward-translated. WMT English$\to$German newstest2019 is a forward-translated test set.}
    \label{fig:testset}
\end{figure}

In addition to the training data, we use newstest2018 for development and newstest2019 for evaluation only. There is an important difference between these two test sets. Newstest2018 was created from monolingual news data
from both English and German online sources. Half of the data 
consists of English text translated into German, while the other half consists of German text translated into English. This results in a joint test set of 2,998 sentences. Newstest2019, on the other hand, consists only of 1,997 sentences translated from English into German (see Figure~\ref{fig:testset}). To provide a joint test set similar to newstest2018, we took newstest2019 from the reverse translation direction German$\to$English, swapped source and target, and concatenated it with the original test sets. This results in a new joint newstest2019 test set of 3,997 sentences.

In addition to reporting overall \BLEU scores on the different test sets, we also report results on the two subsets (based on the original language) of each newstest20XX, which we call the \emph{orig-en} and the \emph{orig-de} halves of the test set.

\newcite{freitag2020bleu} provided an alternative reference translation for the orig-en half of newstest2019. For both standard and alternative references, they
provided an additional paraphrased `as much as possible` version (four different references in all).
In order to enable our parameter tuning experiments, we created a paraphrased version of the reference for the orig-en half of {\em newstest2018} (1,500 sentences) following the instructions from \newcite{freitag2020bleu}. We will release this new paraphrased reference, \emph{newstest2018.orig-en.p}, as part of our work. 
\subsection{Models}

For our translation models, we adopt the transformer implementation from {\it Lingvo}~\cite{shen2019lingvo}, using the transformer-big model size~\cite{vaswani2017attention}. We use a vocabulary of 32k subword units and exponentially moving averaging of checkpoints (EMA decay) with the weight decrease parameter set to $\alpha=0.999$~\cite{buduma2017fundamentals}. We used a batch size of around 32k sentences in all our experiments.

We report \BLEU \cite{papineni2002bleu} in addition to human evaluation. All \BLEU scores are calculated with sacreBLEU \cite{post2018call}\footnote{BLEU+case.mixed+lang.ende+numrefs.1+smooth.exp+
SET+tok.13a+version.1.4.12 SET~$\in$\{wmt18, wmt19, wmt19/google/ar, wmt19/google/arp, wmt19/google/wmtp\}}.

\subsection{Human Evaluation}
\label{sec:huma_eval}

To collect human rankings, we ran side-by-side evaluation for overall quality and fluency. We hired 20 linguists and divided them equally between the two evaluations. Each evaluation included 1,000 items with each item being rated exactly once. We acquired only a single rating per sentence from the professional linguists as we found that they were more reliable than crowd workers \cite{toral2020reassessing}.
We evaluated the orig-en sentences corresponding to the official WMT-19 English$\to$German test set~\cite{barrault2019wmt}. Results in this natural translation direction are more meaningful as pointed out by \citet{Zhang19}, who show that translating a `translationese` source is simpler and should not be used for human evaluation.

\noindent
Our human evaluation followed the protocol:
\begin{itemize}
\item{Fluency:} We present two translations of the same source sentence to professional linguists without showing the actual source sentence. We then ask the rater wether they prefer one of the outputs or rate them equally based on fluency.
\item{Overall Quality:}
We present two translations along with the source and ask 
the raters to evaluate each translation on a 6-point scale. A score of 6 will be assigned to translations with `perfect meaning and grammar`, while a  score of 0 will be assigned to `nonsense/ no meaning preserved` translations. The average over all ratings yields the system's final quality score.
\end{itemize}

\section{Experimental Results}

This section first presents our main result comparing the same system tuned
with BLEU on standard versus paraphrased references. 
We then break down how system design choices impact each metric differently. Throughout, we refer to scores computed with standard references as \BLEU, and those computed with paraphrased references as \BLEUp.

\subsection{Overall Performance}
\label{sec:overall}

We compare the performance of a system optimized on newstest2018 with standard references (opt-on-\BLEU) with one optimized on newstest2018.orig-en with paraphrased references (opt-on-\BLEUp). Both systems were developed using only newstest2018 data, keeping newstest2019 as a blind test set. 
Table~\ref{table:wmt_results_punch} summarizes the results on newstest2019. Details of how these two systems were developed and how they differ are given in Section~\ref{sec:analysing_performance}.

The opt-on-\BLEU system outperforms opt-on-\BLEUp by 5.2 \BLEU points. Normally this would lead us to discard opt-on-\BLEUp. However, the \BLEUp scores tell a different story: opt-on-\BLEUp outperforms by 0.3 points, a potentially large improvement given the smaller natural range of this metric. Under a significance test with random approximation~\cite{riezler2005some}, both the \BLEU and \BLEUp differences are significant at p$<$5e-18. 

\begin{table}[ht]
\begin{center}
{\setlength{\tabcolsep}{.3em}
\begin{tabular}{ l||c|c||}
 & opt-on-\BLEU & opt-on-\BLEUp \\ \hline
 \BLEU & ${\bf{45.0}}$ & $39.8$ \\ \hline
 \BLEUp & $13.4$ & ${\bf{13.7}}$ \\ \hline
 human quality & $4.27$ & ${\bf{4.72}}$ \\ \hline
 human fluency & $27.2\%$ & ${\bf{63.8\%}}$ \\ \hline \hline
\end{tabular}
}
\end{center}
\caption{\BLEU scores and human ratings for WMT newstest2019 English$\to$German (original English sources). We optimized the system to perform best on either newstest2018 with standard reference translations (opt-on-\BLEU) or newstest2018.orig-en with paraphrased reference translations (opt-on-\BLEUp).
BLEU differences are significant according to random approximation \cite{riezler2005some} with p$<$5e-18.
Human score differences are significant according to a Wilcoxon rank-sum test with p$<$5e-18.}
\label{table:wmt_results_punch}
\end{table}

\newcite{freitag2020bleu} showed that BLEU scores calculated on paraphrased references have higher correlation with human judgment than those calculated on standard references. To verify their findings, we ran a human evaluation for the two different outputs on 1,000 sentences randomly drawn from newstest2019 (orig-en), as described above. As shown in Table~\ref{table:wmt_results_punch}, opt-on-\BLEUp is consistently evaluated as better for both quality and fluency. To measure the significance between the two ratings, we ran a Wilcoxon rank sum test on the human ratings and found that both improvements are significant with p$<$e-18.

This experiment demonstrates that we can actually tune our MT system on paraphrased references to yield higher translation quality when compared to a typical system tuned on standard \BLEU. Interestingly, the \BLEU score for the better system is much lower, supporting our contention that \BLEU rewards spurious translation features (e.g. monotonicity and common translations) that are filtered out by \BLEUp.

\subsection{Analysing Performance}
\label{sec:analysing_performance}
We now describe the individual model decisions that went into the two final systems of Section~\ref{sec:overall}. To build a classical system optimized on \BLEU with standard references, we replicate the WMT 2019 winning submission \cite{ng-EtAl:2019:WMT} and examine the effect of each of its major design decisions.\footnote{Our replication achieves 45.0 BLEU on newstest19, competitive with the reference system at 42.7 BLEU.}
In particular, we are looking into the effect of data cleaning, back-translation, fine tuning, ensembling and noisy channel reranking.
We examine the impact of each method on \BLEU and \BLEUp.
For our experiments, we used newstest2018 as our development set and newstest2019 as our held-out test set. All model decisions (checkpoint, variants) are solely made on newstest2018.

Experimental results are presented in Table~\ref{table:wmt_results}. As described in Section~\ref{sec:data}, we report 4 different \BLEU scores for newstest2018 (dev) and newstest2019 (test). In addition to reporting \BLEU score on the joint or the orig-de/orig-en halves of the test sets, we also report \BLEU scores that are calculated on paraphrased references (\BLEUp).

\begin{table*}[ht]
\begin{center}
{\setlength{\tabcolsep}{.27em}
\begin{tabular}{ l||c|c|c|c||c|c|c|c||}
& \multicolumn{4}{c||}{newstest2018 (dev)} & \multicolumn{4}{c||}{newstest2019 (test)} \\ \hline
 & joint & orig-de & orig-en & orig-en.p & joint & orig-de & orig-en & orig-en.p \\ \hline \hline
(1) bitext & 46.0 & 38.8 & 50.6 & 12.8 & 38.5 & 34.9 & 40.9 & 12.1   \\ \hline
(2) + CDS & 46.1 & 39.4 & 50.5 & 13.4 & 39.6 & 35.6 & 42.3 & 12.6 \\ \hline
(3) + BT & 47.2 & 45.3 & 47.7 & 13.6 & 40.9 & 43.1 & 39.4 & 13.1\\ \hline
(4) + Fine tuning & 47.7 & 43.6 & 49.2 & 13.8 & 41.2 & 41.3 & 41.1 & 13.6 \\ \hline
(5) + Ensemble of 4 & 49.8 & 45.4 & 52.1 & 13.7 & 43.1 & 42.1 & 43.6 & 13.3 \\ \hline
+ reranking of (5) (opt on \BLEU) & \bf{50.7} & 44.8 & 53.9 & 13.8 & \bf{43.4} & 41.2 & 45.0 & 13.4 \\ \hline \hline
+ reranking of (4) (opt on \BLEUp)& 47.1 & 45.9 & 47.1 & \bf{14.7} & 41.6 & 44.0 & 39.8 & \bf{13.7} \\ \hline \hline
\end{tabular}
}
\end{center}
\caption{\BLEU scores for WMT 2019 English$\to$German. The {\em joint} sets combine {\em orig-en} and {\em orig-de} subsets. The {\em orig-en.p} sets use paraphrased references instead of standard references. Our experiments compared {\em newstest2018.joint} and {\em newstest2018.orig-en.p} for system tuning. The standard newstest2018 and newstest2019 sets are {\em newstest2018.joint} and {\em newstest2019.orig-en}, respectively.}
\label{table:wmt_results}
\end{table*}

\subsubsection{Data Cleaning}
\label{sec:data_clean}

For data cleaning, we used CDS~\cite{wang-etal-2018-denoising}. We trained a CDS model for English$\to$German taking news-commentary as the in-domain/clean data set. We scored all parallel sentences with our trained CDS model and kept the 70\% highest scoring sentences. Our experimental results suggest that data cleaning is useful for all four types of test sets and consistently improves over a baseline system that is trained on raw parallel data. We conclude that data cleaning is useful for all systems independently of which test set it will be optimized for.

\subsubsection{Back-Translation}
\label{sec:bt}

We trained a strong German$\to$English model on the same parallel data (with flipped source/target) and used that model to (back-)translate (BT) all deduped German monolingual sentences from NewsCrawl 2007-2018 into English. We filtered sentences with a source-target ratio lower than 0.5 or higher than 1.5. We further run language identification and filtered out all backtranslations going into the wrong language. We then oversample our bitext data to match the size of the backtranslation data and train a NMT model on the concatenation of both datasets. 

As previously reported by \cite{Freitag19,bogoychev2019domain}, the original language of the sentences within a test is crucial and can lead to very different conclusions, in particular for back-translation systems. This difference is visible when looking at the \BLEU scores on the standard references. While the \BLEU score on orig-de does improve by 7.5 points, the \BLEU score drops by 2.9 points on the orig-en half. Due to the big gain on the orig-de half, BT also improves the \BLEU score on the joint set. The paraphrased references were designed to overcome these kinds of mismatches and they show a gain of 0.5 BLEU points. We can conclude that back-translation helps improve \BLEU and \BLEUp and we include BT for systems that are optimized for both standard or paraphrased \BLEU scores.

\subsubsection{Fine-Tuning}
\label{sec:fine_tuning}

Similar to \cite{ng-EtAl:2019:WMT}, we fine-tuned our back-translated model on a concatenation of previous WMT testsets (newstest\{2013,2015,2016,2017\}) and the clean in-domain news-commentary corpus. In total, we fine-tuned the model on 330k sentences. We kept all model parameters the same (batch size, learning rate) and continued training on the fine-tuned data for one epoch. 
The \BLEU scores on the standard references suggest a small improvement of 0.3 \BLEU on the joint test set. Interestingly, the improvement is visible on the orig-en half by 0.7 points while the \BLEU scores on orig-de actually drop by 1.7 points. Nevertheless, \BLEUp does improve by 0.5 points, suggesting that fine-tuning is especially helpful when measuring scores with paraphrased references. Despite the small gain on standard references, we include fine-tuning in both our optimized systems.

\subsubsection{Ensemble}

Combining different predictions is a standard approach in MT to boost \BLEU scores. We run ensemble decoding with 4 previously built models. In addition to using the 3 models described in Section~\ref{sec:data_clean}, \ref{sec:bt}, and \ref{sec:fine_tuning}, we build a second fine-tuned model with the same approach, but different initialization.

Although ensemble decoding improves the performance on our standard references by up to 1.9 \BLEU points, the quality is rated as lower by 0.3 \BLEU points on the paraphrased references. We suspect that using an ensemble for decoding favors common, average language by promoting target spans where all systems agree. Paraphrase translations actually downweight the importance of this language, which seems important for agreeing with human judgments \cite{freitag2020bleu}. This promotion of average language and monotonic translation may explain the effectiveness of ensembling only for standard reference \BLEU. Similar to the WMT 2019 winning submission, we include the ensemble approach in our system that is optimized on the joint \BLEU scores. However, we do not include it in our system optimized on \BLEUp.

\subsection{Reranking}

Finally, we extend the noisy-channel approach \cite{yee2019simple} which consists of re-ranking the top-50 beam search output of either the ensemble model (when tuned for \BLEU) or the fine-tuned model (when tuned for \BLEUp). Instead of using 4 features---forward probability, backward probability, language model and word penalty---we use 11 forward probabilities, 10 backward probabilities and 2 language model scores. Different to \cite{ng-EtAl:2019:WMT}, we did not pick the re-ranking weights through random search, but used MERT~\cite{och2003minimum} for efficient tuning. 

The 11 different forward translation scores come from different English$\to$German NMT models that are replicas of the previous described models (Section~\ref{sec:data_clean}, \ref{sec:bt}, and \ref{sec:fine_tuning}). The 10 backward translation scores come from the same approaches, but trained in the reverse direction. These 21 NMT model scores are combined with 2 language model (LM) scores. The first LM is trained on the German monolingual NewsCrawl data, while the second LM is trained on forward-translated English NewsCrawl data. The first LM should assign high scores to genuine German text, while the second LM should assign high scores to translationese German originating from English.

We first reranked the 50-best list generated by the ensemble model with MERT on newstest2018. Similar to the original WMT 2019 submission, the \BLEU scores on the joint and orig-en set increase. This reranked output corresponds to our opt-on-\BLEU model.
Next, we reranked the 50-best list generated by the fine-tuned model with MERT on newstest2018.orig-en with paraphrased references. This led to further small increases in \BLEUp, and corresponds to our opt-on-\BLEUp model.

In summary, optimizing on \BLEUp leads us to keep back-translation, even though evaluation with standard English-original references would have us drop it, and also leads us to drop the ensembling step. Rescoring using MERT weights learned with \BLEU or \BLEUp further separates the systems according to these metrics.

\section{Analysis}

This section confirms the results from the previous section with additional references for newstest2019 and illustrates the behaviour of our systems on individual sentences.

\subsection{Alternative Reference Translations}

\newcite{freitag2020bleu} released an additional standard reference translation (AR) and two `paraphrase as-much-as-possible` reference translations for newstest2019 (WMT.p and AR.p). We used WMT.p in all our above experiments; here we report \BLEU scores for all four available reference translations in table~\ref{table:wmt_results_ar}. The \BLEU improvements between the two standard reference translations agree perfectly. Similarly, the \BLEUp improvements between the two paraphrased references also coincide. 
This indicates that by optimizing on \BLEU or \BLEUp we have not somehow overfit to a specific set of reference translations or their paraphrases, but instead have molded our model to better match a style of reference translation.

\begin{table*}[ht]
\begin{center}
\begin{tabular}{ l|||c|c|c|c||}
& \multicolumn{4}{c||}{newstest2019} \\ \hline
& WMT & AR & WMT.p & AR.p \\
& (orig-en) & (orig-en) & (orig-en.p) & (orig-en.p) \\ \hline \hline
(1) bitext  & 40.9 & 32.2 & 12.1 & 12.0 \\ \hline
(2) + CDS  & 42.3 & 34.2 & 12.6 & 12.3 \\ \hline
(3) + BT  & 39.4 & 33.6 & 13.1 & 13.0\\ \hline
(4) + Fine tuning  & 41.1 & 35.5 & 13.6 & 13.4 \\ \hline
(5) + Ensemble of 4  & 43.6 & 36.0 & 13.3 & 13.0 \\ \hline
+ reranking of (5) (opt-on-\BLEU) & \bf{45.0} & \bf{36.7} & 13.4 & 13.1 \\ \hline \hline
+ reranking of (4) (opt-on-\BLEUp)& 39.8 & 34.4 & \bf{13.7} & \bf{13.5} \\ \hline  \hline
\end{tabular}
\end{center}
\caption{\BLEU scores for English$\to$German newstest2019 for the additional references from \cite{freitag2020bleu}.}
\label{table:wmt_results_ar}
\end{table*}

\subsection{Translation Examples}

This section presents translation examples from our two differently optimized systems in Table~\ref{tab:example_output}. 
The first 3 examples show sentences where opt-on-\BLEUp has higher translation quality than opt-on-\BLEU.
One observation of \cite{freitag2020bleu} was that \BLEU scores calculated on standard references prefer monotonic translations. This is visible in our first translation example, where opt-on-\BLEU incorrectly translates the saying {\em Tomorrow's a different beast} into {\em Morgen ist ein anderes Biest}, using an inappropriately monotonic strategy. On the other hand, the opt-on-\BLEUp system captures the meaning of the source sentence and generates a valid translation.

Another drawback of standard reference \BLEU is the preference for literal translation. This is visible in our second example where the word {\em cap} is translated into {\em Kappe} and {\em tip} into {\em kippen}. Both are valid word-by-word translations, but do not make much sense in this context. The third example is another example of the monotonic translation style of a regular tuned system. The opt-on-\BLEU translation is an incorrect word-by-word translation. The opt-on-\BLEUp system is able to introduce a German natural sentence structure and generate a flawless translation.

The last translation example is a loss for the paraphrased-tuned system and demonstrates that sometimes a more literal translation can be better. Even though the word {\em run} can be translated into {\em Ansturm}, it is not appropriate in this context and the simpler translation {\em Lauf} is correct.

\begin{table*}[ht]
    \centering
    \setlength\tabcolsep{4pt}
    \begin{tabular}{c|l}
    \hline
    source & Tomorrow's a different beast. \\
    opt on \BLEU & Morgen ist ein anderes Biest. \\
    opt on \BLEUp & Morgen ist alles anders. \\ \hline
    source & You have to tip your cap. \\
    opt on \BLEU & Sie müssen Ihre Kappe kippen. \\
    opt on \BLEUp & Man muss den Hut ziehen. \\ \hline
    source & He averaged 5.6 points and 2.6 rebounds a game last season. \\
    opt on \BLEU & Er durchschnittlich 5,6 Punkte und 2,6 Rebounds ein Spiel in der vergangenen Saison. \\
    opt on \BLEUp & In der vergangenen Saison erzielte er im Schnitt 5,6 Punkte und 2,6 Rebounds pro Spiel. \\ \hline
    source & Thirty-two percent supported such a run. \\
    opt on \BLEU & 32 Prozent unterst\"utzten einen solchen Lauf. \\
    opt on \BLEUp & 32 Prozent sprachen sich für einen solchen Ansturm aus. \\ \hline

 \hline
    \end{tabular}
    \caption{Example output for English$\to$German for  systems optimized on standard \BLEU or \BLEUp. Translations for opt-on-\BLEU tend to be more literal, and adhere closely to the source sentence structure.}
\label{tab:example_output}
\end{table*}

\subsection{Matched n-grams}
The \BLEU scores calculated on the two different references yield different conclusions. \BLEU on standard references evaluated opt-on-\BLEU higher by more than 5 \BLEU points. \BLEUp came to a different conclusion and gave a higher score to opt-on-\BLEUp. In this section, we look at the n-grams that contributed most to these different outcomes.
Those that contribute most to the difference in \BLEU across the two systems are:
\begin{itemize}
    \setlength\itemsep{-0.2em}
    \item \textbf{Er sagte, dass} (\textit{He said that})
    \item \textbf{, sagte er der} (\textit{, he said the})
    \item \textbf{stellte fest, dass} (\textit{noted that})
\end{itemize}
These are all generic, high-frequency n-grams. They are crucial for attaining high BLEU scores, and tend to appear in translations that employ the same structure as the source sentence.
In contrast, the n-grams that contribute most to the difference in \BLEUp are:
\begin{itemize}
    \setlength\itemsep{-0.2em}
    \item \textbf{Menschen ums Leben kamen} (\textit{humans died})
    \item \textbf{Grossbritanien keine Steuern zahlen} (\textit{Great Britain pay no tax})
    \item \textbf{von BBC Scottland} (\textit{from BBC Scottland})
\end{itemize}
These are much less frequent sequences with more semantic content.

\section{Conclusions}

Prior work has shown that BLEU measured on paraphrased references (\BLEUp) has better correlation with human evaluation than BLEU measured on regular references (\BLEU) for the comparison of {\it existing} systems~\cite{Freitag19}. Motivated by this finding, we collected a development set of paraphrased references and assessed \BLEUp for {\it system development}. This allowed us to evaluate if the design choices of a modern neural MT system impact \BLEU and \BLEUp differently, including tuning a re-ranking noisy channel model to these metrics. Our experiments followed the setup from the winning newstest19 English$\to$Germam entry at WMT19~\cite{ng-EtAl:2019:WMT}.

For design choices, we observe that \BLEUp seems to emphasize the importance of back-translation even when test sets are source original. On the other end, \BLEUp seems to de-emphasize the importance of ensembles, as the reliable prediction of common language by ensembles is less rewarded by this metric.

Our tuning experiments led to positive results. In human evaluation, the system tuned on \BLEUp showed significant improvements in terms of adequacy and even greater gains in terms of fluency compared to the system tuned on \BLEU. Example translations indicate that the model tuned on \BLEUp produces noticeably less literal translations. Our experiments also highlight a disconnect between regular \BLEU and human evaluation: the system tuned on \BLEUp degrades standard \BLEU scores by over 5 points, while faring significantly better in human evaluation. Paraphrased automatic evaluation therefore seems to be a promising proxy for human evaluation when making design choices for MT systems.

This research opens the question of whether these results can be confirmed over a wide range of language pairs. We also hope to achieve further improvements by refining the paraphrased evaluation protocol.

\bibliographystyle{acl_natbib}
\bibliography{anthology,emnlp2020}

\end{document}